\documentclass[]{amsart} 

\usepackage{graphicx}
\usepackage[T1]{fontenc} 
\usepackage{ae, aecompl} 
\usepackage{booktabs}
\usepackage{fix-cm}

\usepackage{amssymb}
\usepackage{amsmath}
\usepackage{longtable,verbatim}
\usepackage{booktabs}
\usepackage{color, colortbl}
\usepackage{algorithm}
\usepackage{algorithmic}
\usepackage{subfigure}
\usepackage{url}
\begin{document}

\title[Generative Adversarial Networks in EDAs]{Generative Adversarial Networks in Estimation of Distribution Algorithms for Combinatorial Optimization}

\author[]{Malte Probst}
\address[]{Johannes Gutenberg-Universität Mainz\\
Dept.~of Information Systems and Business Administration\\
Jakob-Welder-Weg 9, 55128 Mainz, Germany}
\email{probst@uni-mainz.de}
\urladdr{http://wi.bwl.uni-mainz.de}
\begin{abstract}
Estimation of Distribution Algorithms (EDAs) require flexible probability models that can be efficiently learned and sampled. Generative Adversarial Networks (GAN) are generative neural networks which can be trained to implicitly model the probability distribution of given data, and it is possible to sample this distribution. We integrate a GAN into an EDA and evaluate the performance of this system when solving combinatorial optimization problems with a single objective. We use several standard benchmark problems and compare the results to state-of-the-art multivariate EDAs. GAN-EDA doe not yield competitive results -- the GAN lacks the ability to quickly learn a good approximation of the probability distribution. A key reason seems to be the large amount of noise present in the first EDA generations.
\end{abstract}


\maketitle

\section{Introduction}
\label{intro}
Estimation of Distribution Algorithms (EDA) \cite{Muehlenbein1996,larranaga2002estimation} are metaheuristics for combinatorial and continuous non-linear optimization. The maintain a population of candidate solutions, and iteratively improve the quality of this population over multiple generations. In each generation, they first select solutions with high quality from the population. Subsequently, they build a model that approximates the probability distribution of these solutions. Then, new candidate solutions are sampled from the model. The EDA then starts over by selecting the next set of good solutions from the new candidate solutions and the previous selection.

In order to be suitable for an EDA, a probabilistic model has to fulfill certain criteria:
\begin{itemize}
 \item It must approximate the probability distribution of the selected individuals with sufficient quality (either explicitly or implicitly). 
 \item It must be able to sample new solutions from this probability distribution, which the EDA can use as candidate solutions for the next generation
 \item Both learning and sampling must be computationally efficient, i.e., fast
\end{itemize}    

Previous work has shown that generative neural networks can lead to competitive performance. \cite{Probst2014} use a Restricted Boltzmann Machine (RBM) in an EDA and show that RBM-EDA can achieve competitive performance to state-of-the art EDAs, especially in terms of computational complexity of the CPU time. \cite{probst2015dbm} propose DBM-EDA, where the probabilistic model is a Deep Boltzmann Machine (DBM). DBM-EDA shows competitive performance on problems which can be decomposed into independent subproblems, but inferior performance on other problem types.
\cite{probst2015dae-eda-better-arxiv} use a Denoising Autoencoder (DAE) in an EDA. DAE-EDA achieves superior performance for problems which can be factorized, i.e., decomposed into independent subproblems. Furthermore, the computational effort of DAE-EDA is low, compared to other EDAs.
In general, neural-network inspired probabilistic models can often be parallelized on massively parallel systems such as graphics processing units (GPUs) \cite{Probst2014a}.

In this paper, we focus on Generative Adversarial Networks (GAN) \cite{goodfellow2014generative}. GANs are generative neural networks that implicitly estimate the probability distribution of given data using two adversaries - the generator $G$ and the discriminator $D$. $G$'s task is to produce samples with the desired properties, $D$ tries to distinguish $G$'s samples from real samples. $G$ and $D$ are trained iteratively.  

We implemented GAN-EDA and tested its performance on the onemax problem, concatenated trap functions, NK landscapes and the HIFF problem. We compare the results the state-of-the-art multivariate Bayesian Optimization Algorithm (BOA, see \cite{Pelikan1999,Pelikan2005}), and DAE-EDA. We publish the source code of all experiments\footnote{See https://github.com/wohnjayne/eda-suite/ for the complete source code}.
The paper is structured as follows: In Section \ref{gan}, we introduce GANs. Section \ref{setup} briefly describes test problems, comparison algorithms, and the experimental setup. The results are presented in Section \ref{results}.
\section{Generative Adversarial Nets}
\label{gan}
\begin{figure}
\begin{center}
\centerline{\includegraphics[width=0.4\columnwidth]{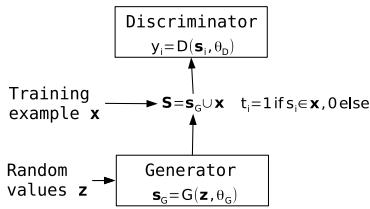}}
\caption{A schematic view of a Generative Adversarial Network. The generator $G$ takes random values $\mathbf{z}$ as input, and generates samples $\mathbf{s}_G$. The Discriminator $D$ takes sample $\mathbf{s}_i$ as input and calculates the probability $y_i$ that $\mathbf{s}_i$ came from the training data $x$. $G$'s objective is to maximize $D(\mathbf{s}_G)$, while $D$ tries to maximize its discriminative performance between $\mathbf{x}$ and $\mathbf{s}_G$. $G(\cdot)$ and $D(\cdot)$ can be arbitrary differentiable functions.}
\label{fig-dbm}
\end{center}
\end{figure}
\begin{algorithm}
\caption{Pseudo code for training a GAN}
\label{alg-gan}
\begin{algorithmic}[1]
\STATE \textbf{Set} $0<\alpha<1$, e.g. $\alpha=0.1$
\STATE \textbf{Initialize} $\theta_G,\theta_D$ to small random values
\WHILE {not converged}
    \FOR{\textbf{each} example $\mathbf{x}$ in the training set}
        \STATE \textit{--- Discriminator Training  ---}
        
        \STATE $\mathbf{z} \leftarrow$ sample a random $\mathbf{z}$
        \STATE $\mathbf{s}_G \leftarrow$ create a generator sample $\mathbf{s}_G=G(\mathbf{z},\theta_G)$
        \FOR{\textbf{each} sample $\mathbf{s}_i$ in $\{\mathbf{s}_G,\mathbf{x}\}$}
            \STATE $y_i=D(\mathbf{s}_i,\theta_D)$
            \STATE $\theta_D:=\theta_D-\alpha*\frac{\partial l_D(y_i,t_i)}{\partial\theta_D}$,
            \STATE $\quad$ with $t_i=1 \text{ if } \mathbf{s}_i \text{ is a training example, 0 else}$
        \ENDFOR
        \STATE \textit{--- Generator Training  ---}
        \STATE $\mathbf{z} \leftarrow$ sample a random $\mathbf{z}$
        \STATE $\mathbf{s}_G \leftarrow$ create a generator sample $\mathbf{s}_G=G(\mathbf{z},\theta_G)$
        \STATE $y_i=D(\mathbf{s}_i,\theta_D)$
        \STATE $\theta_G:=\theta_G-\alpha*\frac{\partial l_G(z)}{\partial\theta_D}$,
        \STATE $\quad$ using error backpropagated by $D$
    \ENDFOR
\ENDWHILE
\end{algorithmic}
\end{algorithm}
Generative Adversarial Nets \cite{goodfellow2014generative} use two components to estimate the probability distribution of given data\footnote{We use the following notation:   $x$ denotes a scalar value, $\mathbf{x}$ denotes a vector of scalars, $\mathbf{X}$ denotes a matrix of scalars} $\mathbf{x}\in R^n$. The generator, $G$, takes random values $\mathbf{z}$ from some prior probability distribution $p(\mathbf{z})$ as input and maps them to an output $\mathbf{s}_G=G(\mathbf{z};\theta_G)$ of size $n$. $\theta_G$ are $G$'s learnable parameters.
Then, let $\mathbf{S}$ be the union of all $m$ examples in the training set and $m$ of $G$'s samples.
The discriminator $D$ takes samples $\mathbf{s}_i$ from $\mathbf{S}$ as input, with $i=1\dots 2*m$.
Thus, $\mathbf{s}_i$ can be either a sample $G(\mathbf{z})$ generated by $G$, or a "real" sample $\mathbf{x}$ from the data.
$D$ then maps $\mathbf{s}$ to a scalar value $y=D(\mathbf{s};\theta_D)\in(0,1)$, which is interpreted as the probability that $\mathbf{s}$ comes from the real data instead of $G$.  $\theta_D$ are $D$'s learnable parameters. Hence, the goal of $D$ is to discriminate well between samples from $G$ and real data, and $G$'s objective is to fool $D$. That is, $G$'s goal is to produce samples which come from the same probability distribution as $\mathbf{x}$.

Accordingly, $D$ is trained to maximize its discriminative performance (see Algorithm \ref{alg-gan}). It tries to find parameters $\theta_D^*$ to minimize the cross-entropy loss $l_D(y_i,t_i)$ for all examples $\mathbf{s}_i$, where, by definition, $t_i=1$ if $s_i$ is an example from the training set, and $t_i=0$ if $s_i$ was generated by $G$. Therefore,
\begin{equation}
\label{eq-lossD}
\theta_D^*=\underset{\theta_D}{\text{argmin}}-\sum_{i=1}^{2*m}{[t_i*\log(y_i) + (1-t_i)*\log(1-y_i)]}
\end{equation}
Finding $\theta_D^*$ is performed by stochastic gradient descent. That is, for each example $i$, $\theta_D$ is updated in the direction of the gradient $\frac{\partial l(y_i,t_i)}{\partial \theta_D}$, using the backpropagation algorithm.

The generator $G$ is optimized using gradient descent as well. Recall that its objective is to fool $D$. Hence, it tries to convince $D$ that $G$'s samples are proper training examples by maximizing the second term in the sum of Equation \ref{eq-lossD}. It therefore tries to minimize the loss function $l_G(\mathbf{z})=\log(1-D(G(\mathbf{z})))$, as
\begin{equation}
\label{eq-lossG}
\theta_G^*=\underset{\theta_G}{\text{argmin}}\sum_{i=1}^{m}{\log(1-D(G(\mathbf{z}_i)))},
\end{equation}
for all examples $\mathbf{z}_i, i=1\dots m$ generated by $G$. This, again, is performed using stochastic gradient descent, and backpropagation. Note that, in order to train $G$, the errors are backpropagated through $D$ first. In other words, $G$ is being told how to change its output such that $D$ will accept the fraud as genuine.

In principle, $G(\cdot)$ and $D(\cdot)$ can be arbitrary differentiable functions. Here, we use multi layer percerptrons, i.e., feed-forward neural networks (see Section \ref{test-setup}).
\section{Experimental Setup}
\label{setup}
\subsection{Test Problems}
We evaluate DAE-EDA on onemax, concatenated deceptive traps \cite{deb1993analyzing}, NK landscapes \cite{kauffman1989nk} and the HIFF function \cite{watson1998modeling}. All four are standard benchmark problems. Their difficulty depends on the problem size, i.e., problems with more decision variables are more difficult. Furthermore, the difficulty of concatenated deceptive trap functions and NK landscapes is tunable by a parameter. Apart from the simple onemax problem, all problems are composed of subproblems, which are either deceptive (traps), overlapping (NK landscapes), or hierarchical (HIFF), and therefore multimodal.  We use instances of NK landscapes with known optima from \cite{Pelikan2008techreport2}.
\subsection{Algorithms for Comparison}
We compare the performance of GAN-EDA to two other EDAs, BOA and DAE-EDA.
BOA is a state-of-the-art multivariate EDA \cite{Pelikan1999,Pelikan2005}. It uses a Bayesian network to represent dependencies between the decision variables. It learns the structure of the network using a greedy algorithm, which iteratively adds edges to the network. BOA's probabilistic model is usually very accurate, leading to a low number of fitness evaluations. However, the computational effort for the greedy search algorithm is high \cite{Probst2014}.

DAE-EDA uses a Denoising Autoencoder (DAE) as its probabilistic model  \cite{vincent2008extracting,probst2015dae-eda-arxiv,probst2015dae-eda-gecco}. The DAE is a feed-forward neural network, which is trained using the backpropagation algorithm. Its objective is reconstruct the data presented to the input. Between input and output layer, DAEs have one or multiple hidden layers. Using a special sampling procedure, a trained DAE can sample from the approximated distribution of the training data \cite{Bengio-et-al-NIPS2013}. Note that the results for DAE-EDA shown in this paper come from the updated version of the algorithm, which yields better results than the initial version (publication of the new version is under preparation \cite{probst2015dae-eda-better-arxiv}).
\subsection{Test Setup}
\label{test-setup}
We exemplarily show the performance of GAN-EDA on four different instances, one of each test problem (see Section \ref{results}). For each algorithm and instance, we test multiple population sizes. For each population size, we perform 20 independent runs. We report the average fitness of these runs, in relation to the population size, the number of unique fitness evaluations, and the required CPU time.

We implement $G(\cdot)$ and $D(\cdot)$ as feed-forward neural networks. We performed a number of preliminary tests with different architectures and hyper-parameters, thereby exploring different settings for
\begin{itemize}
    \item the topology, i.e., number and size of hidden layers,
    \item the types of neurons (i.e., activation functions), i.e.,
    \begin{itemize}
        \item sigmoid units, with $y=\frac{1}{1+exp(-x)}$
        \item rectified linear units, with $y=\text{max}(0,x)$
        \item maxout units  \cite{goodfellow2013maxout}
    \end{itemize}
    \item the types of regularization, i.e.,
    \begin{itemize}
        \item dropout regularization \cite{srivastava2014dropout}
        \item L2 regularization (weight decay)
    \end{itemize}
    \item the use of \textit{momentum} \cite{qian1999momentum},
    \item the details of the initialization of the parameters $\theta_G$ and $\theta_D$ (normal or uniform distribution),
    \item other hyper-parameters such as schedules for the learning rate.
\end{itemize}
For the results presented in Section \ref{results}, we used networks with one hidden layer of rectified linear units, both for $G$ and for $D$. All hyper-parameters, and further details of the learning process such as schedules for momentum and weight decay are saved in a configuration file along with the source code (see git repository on github.com). 

The algorithms are implemented in Matlab/Octave and executed using Octave V3.2.4 on a on a single core of an AMD Opteron 6272 processor with 2,100 MHz.  
\section{Results and Discussion}
\label{results}
\begin{figure}
        \subfigure
        {
                \includegraphics[width=0.34\linewidth,angle=270]{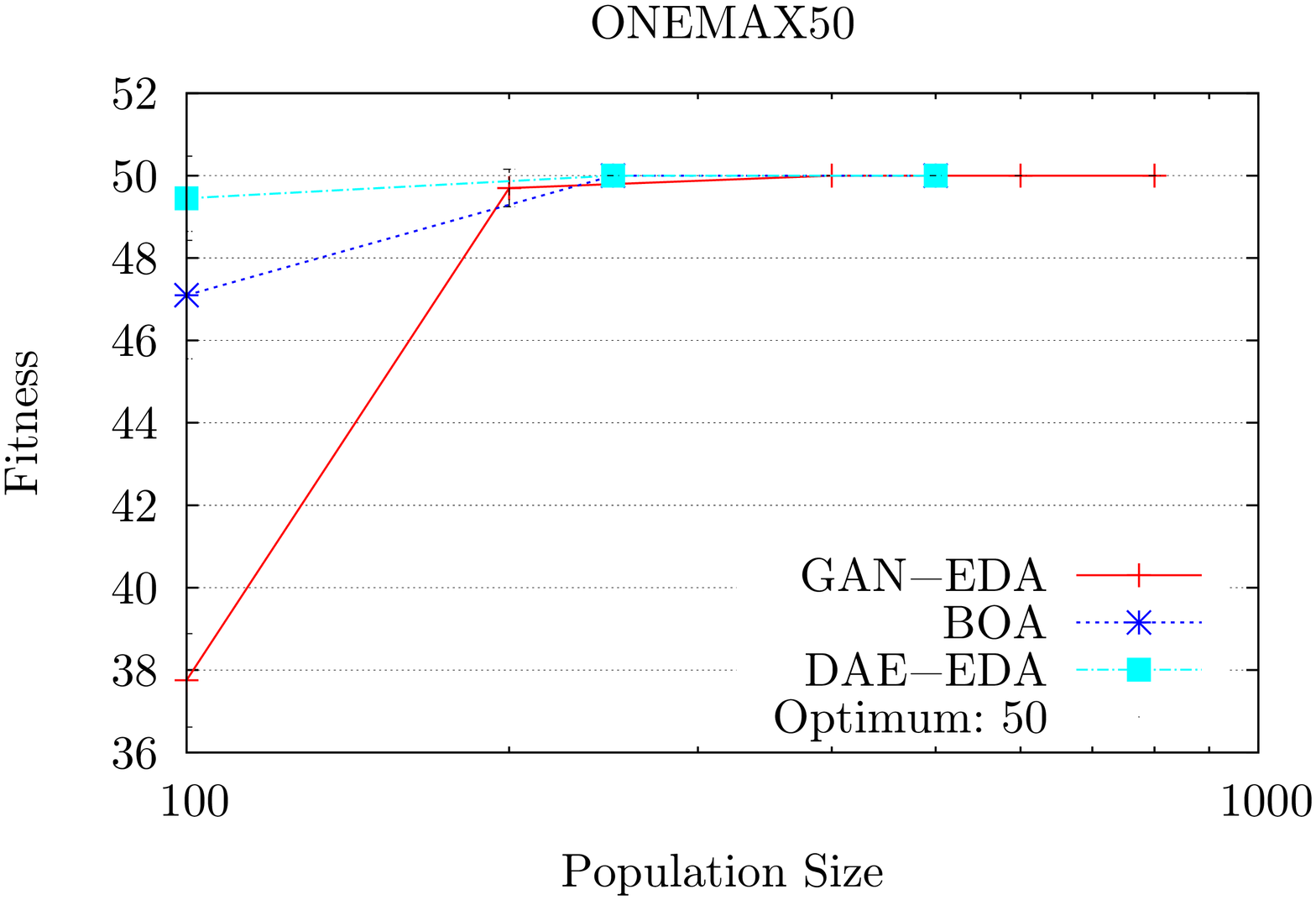}
                \includegraphics[width=0.34\linewidth,angle=270]{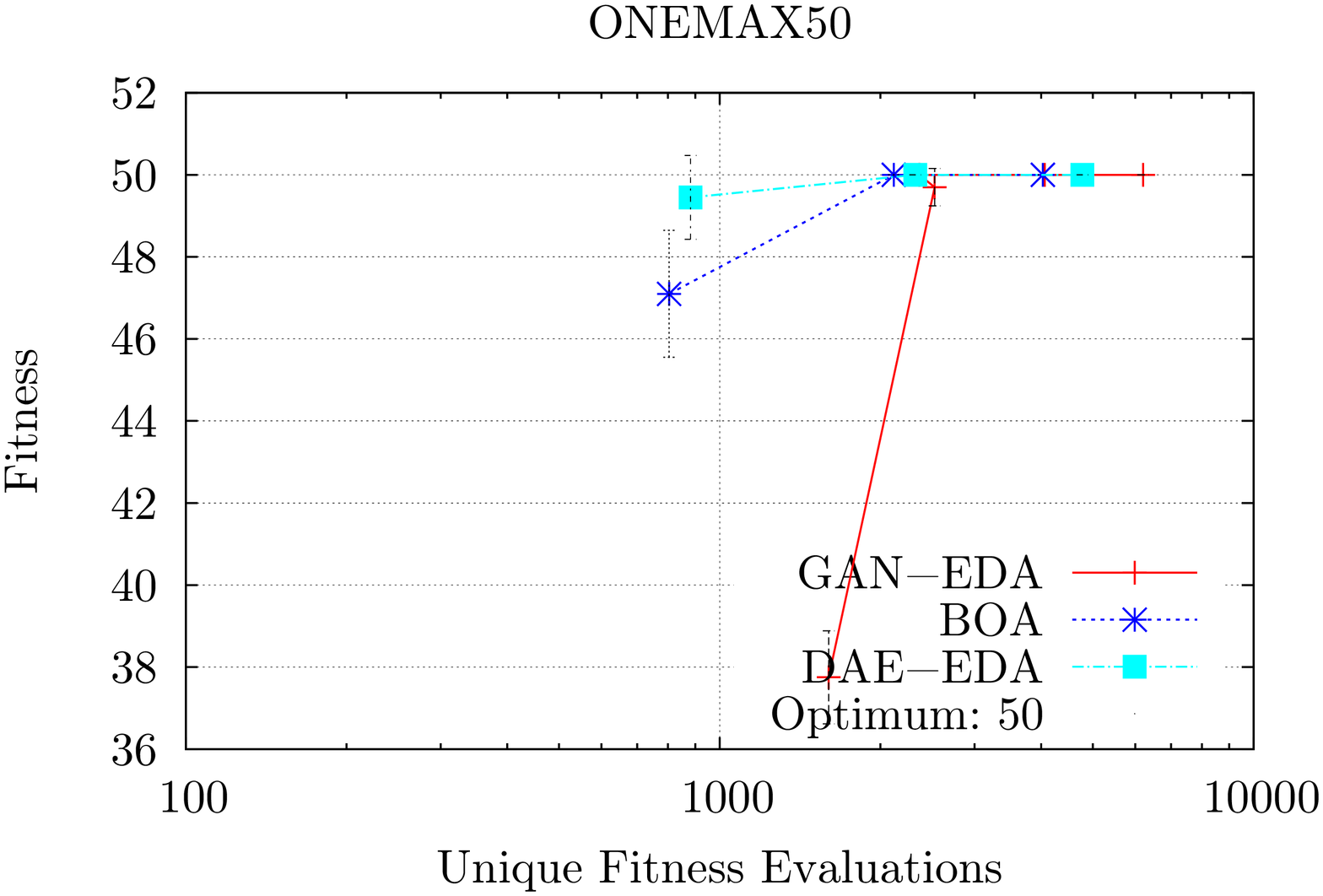}
        }
        \subfigure
        {
                \includegraphics[width=0.34\linewidth,angle=270]{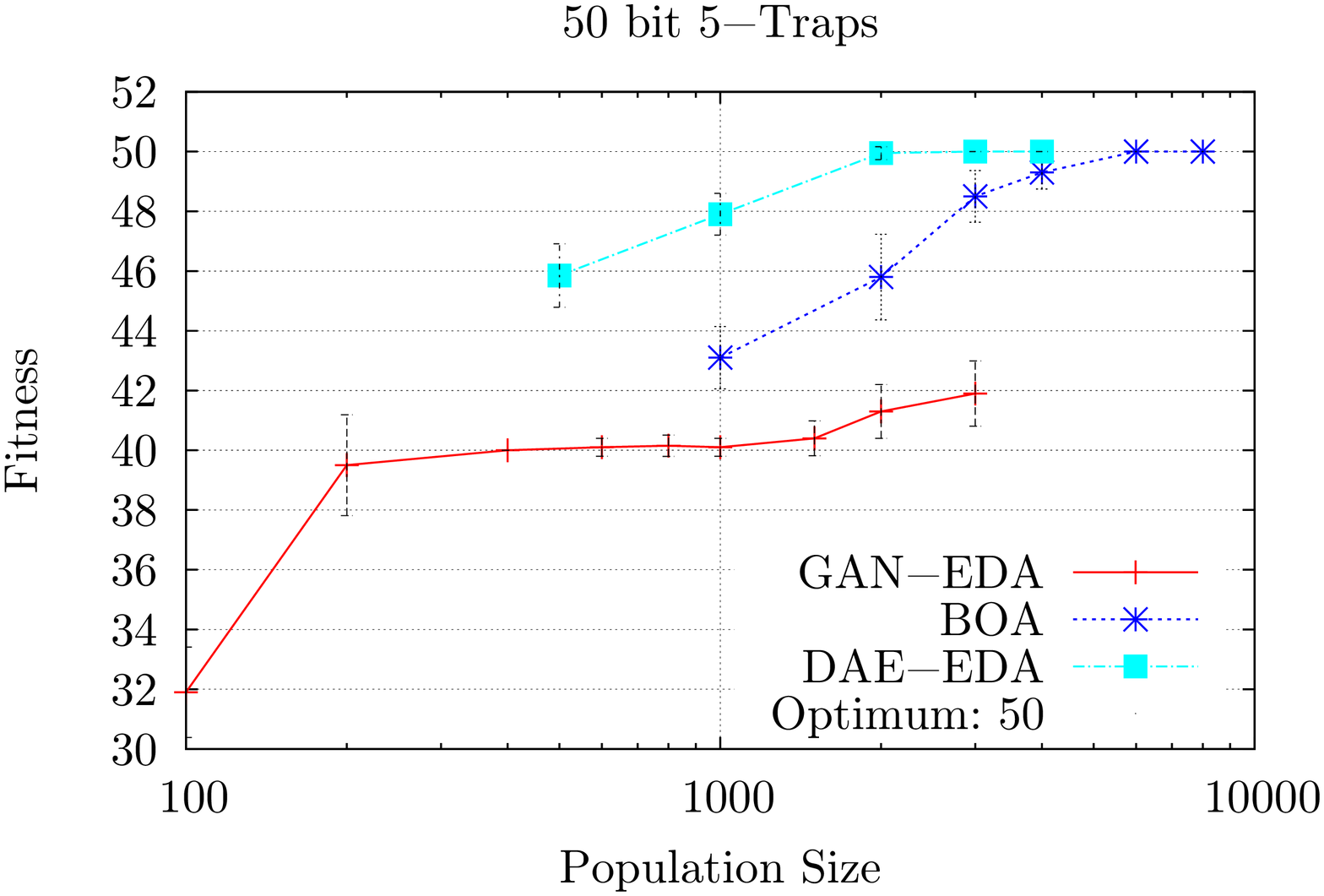}
                \includegraphics[width=0.34\linewidth,angle=270]{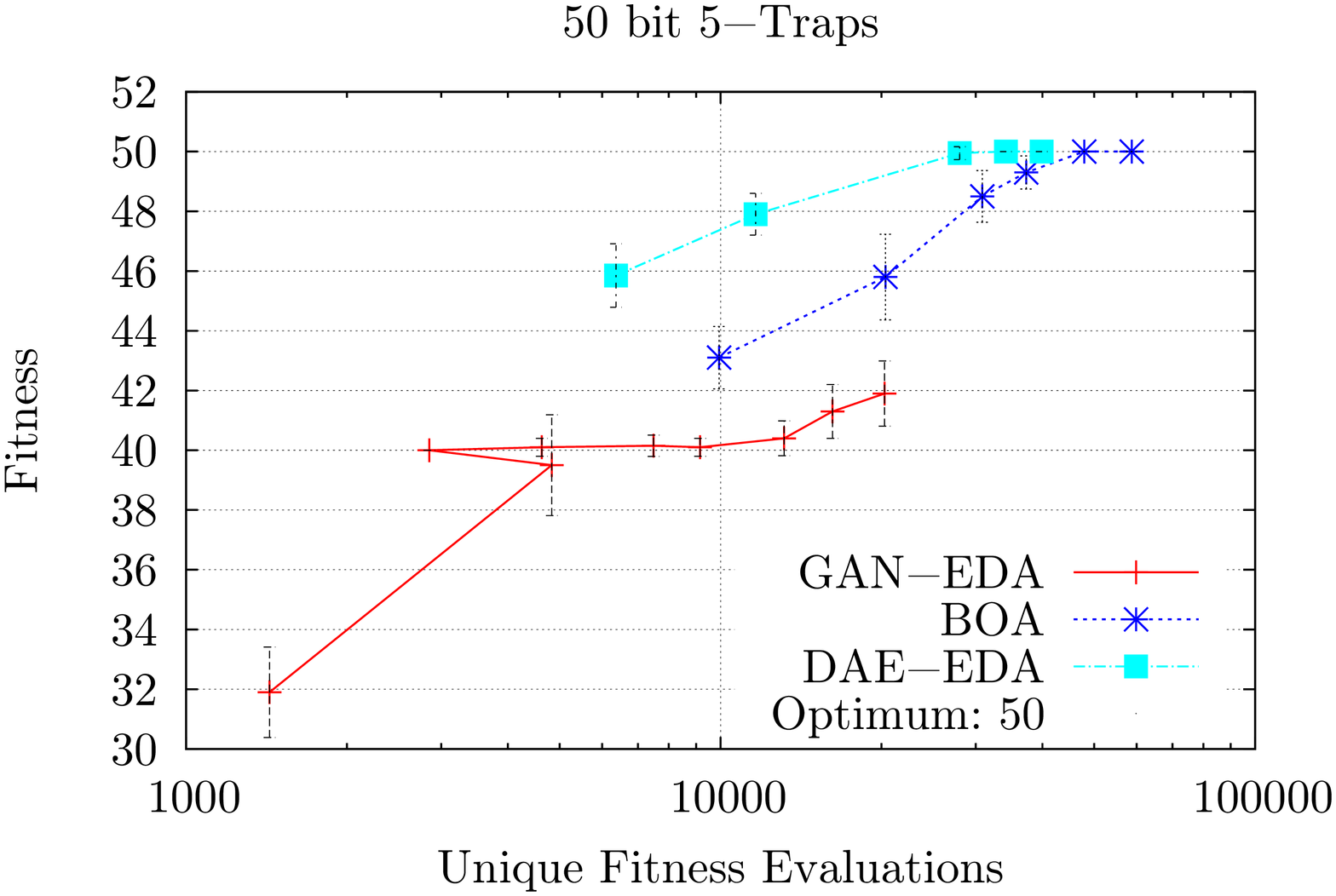}
        }
        \subfigure
        {
            \includegraphics[width=0.34\linewidth,angle=270]{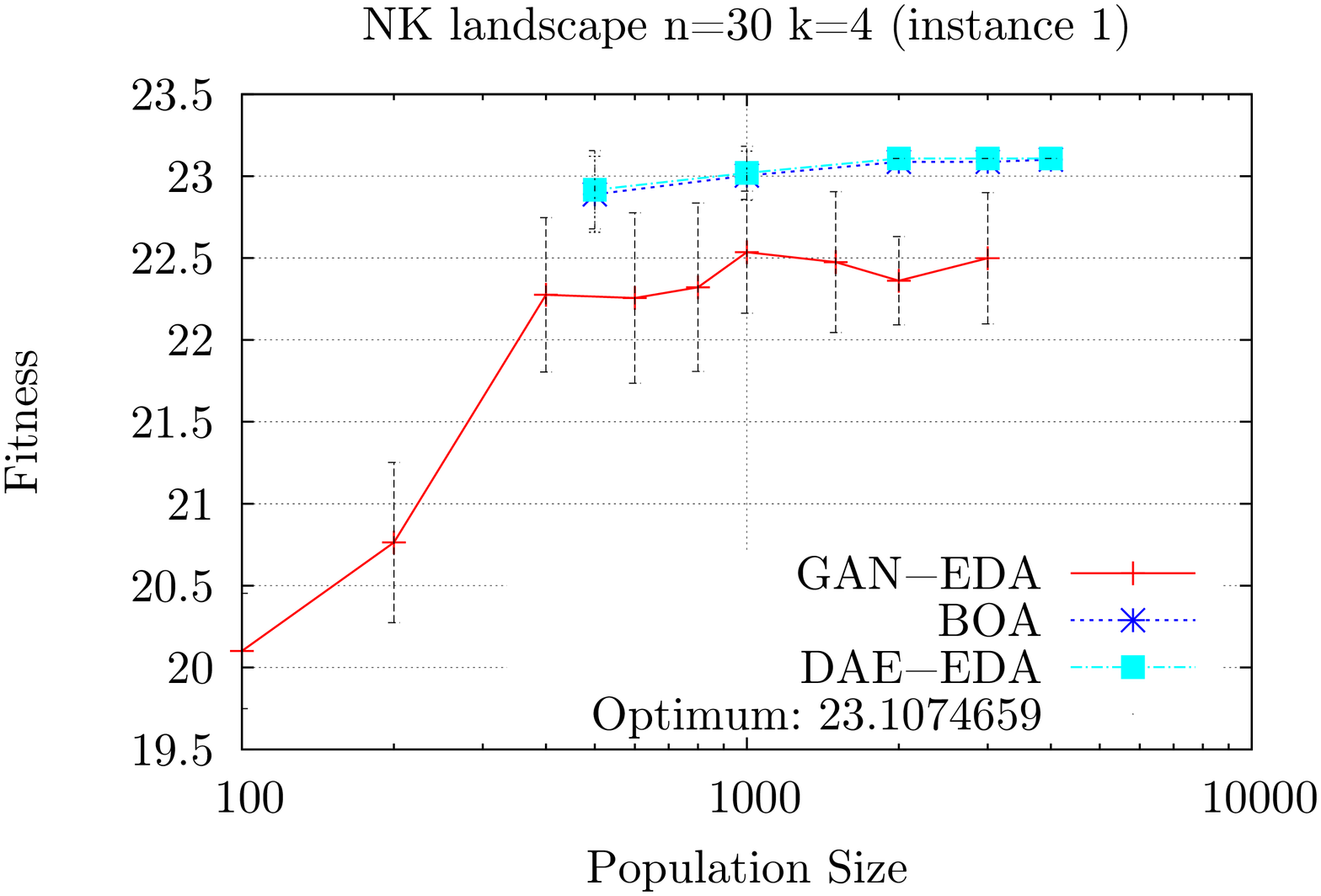}
            \includegraphics[width=0.34\linewidth,angle=270]{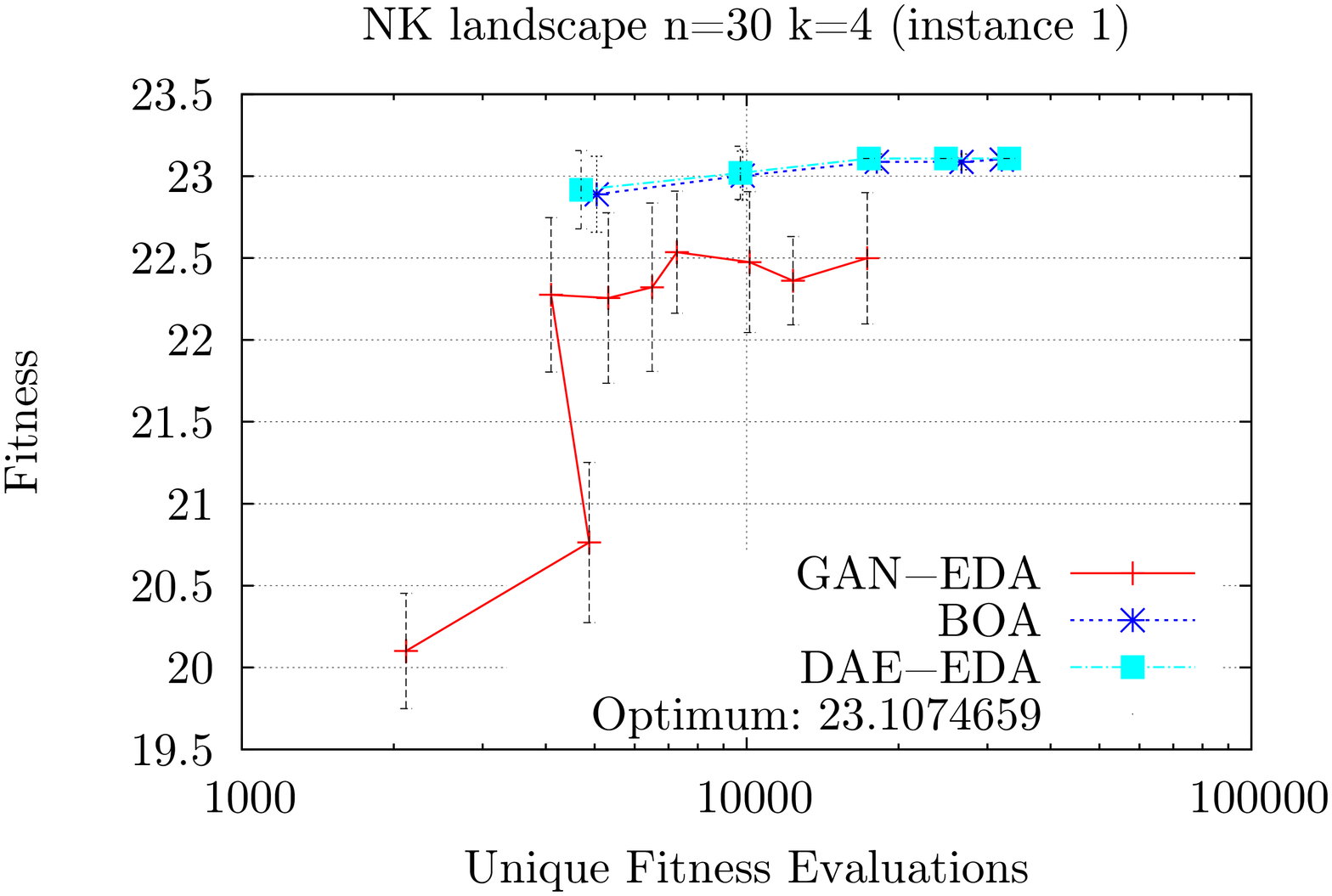}
        }
        \subfigure
        {
            \includegraphics[width=0.34\linewidth,angle=270]{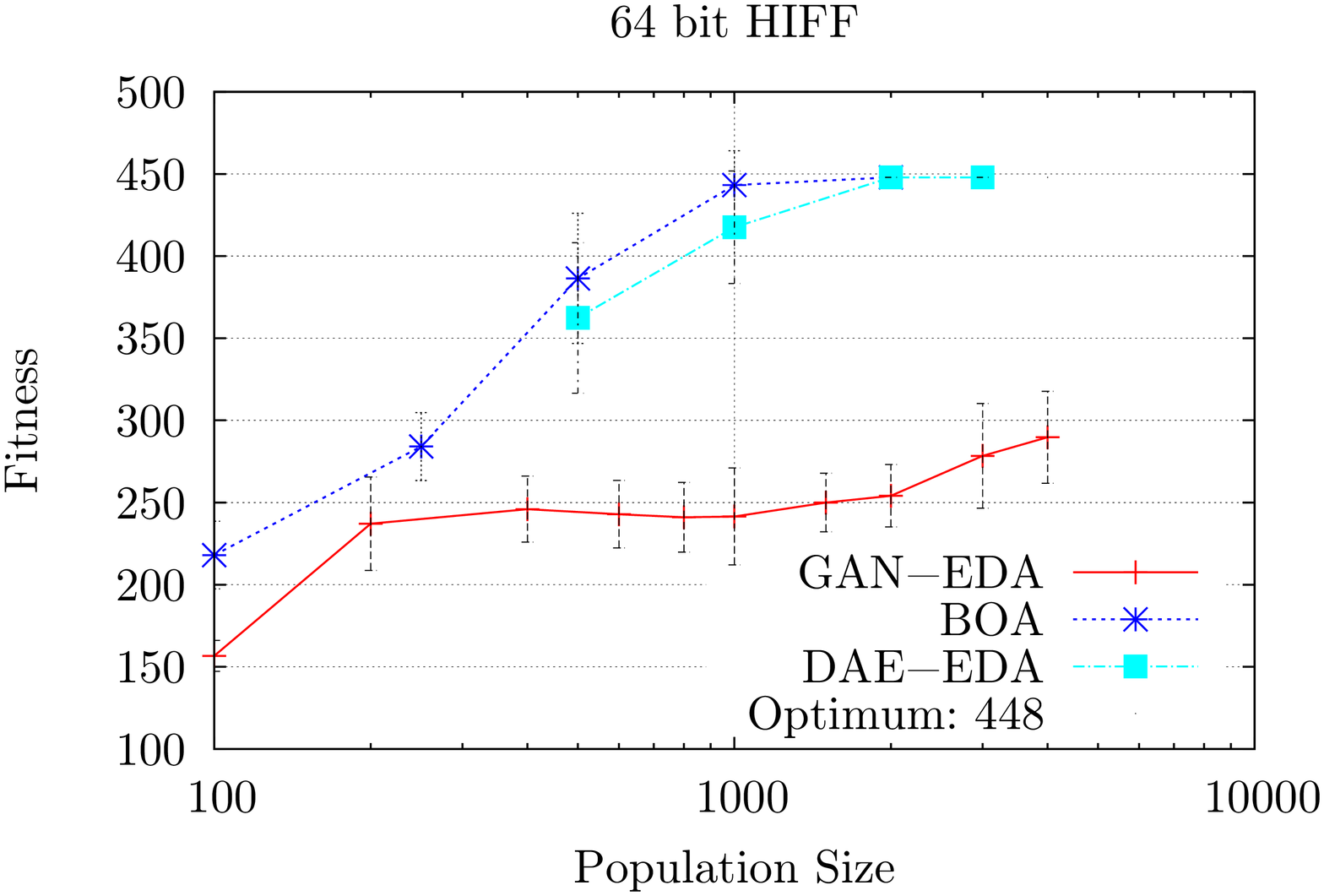}
            \includegraphics[width=0.34\linewidth,angle=270]{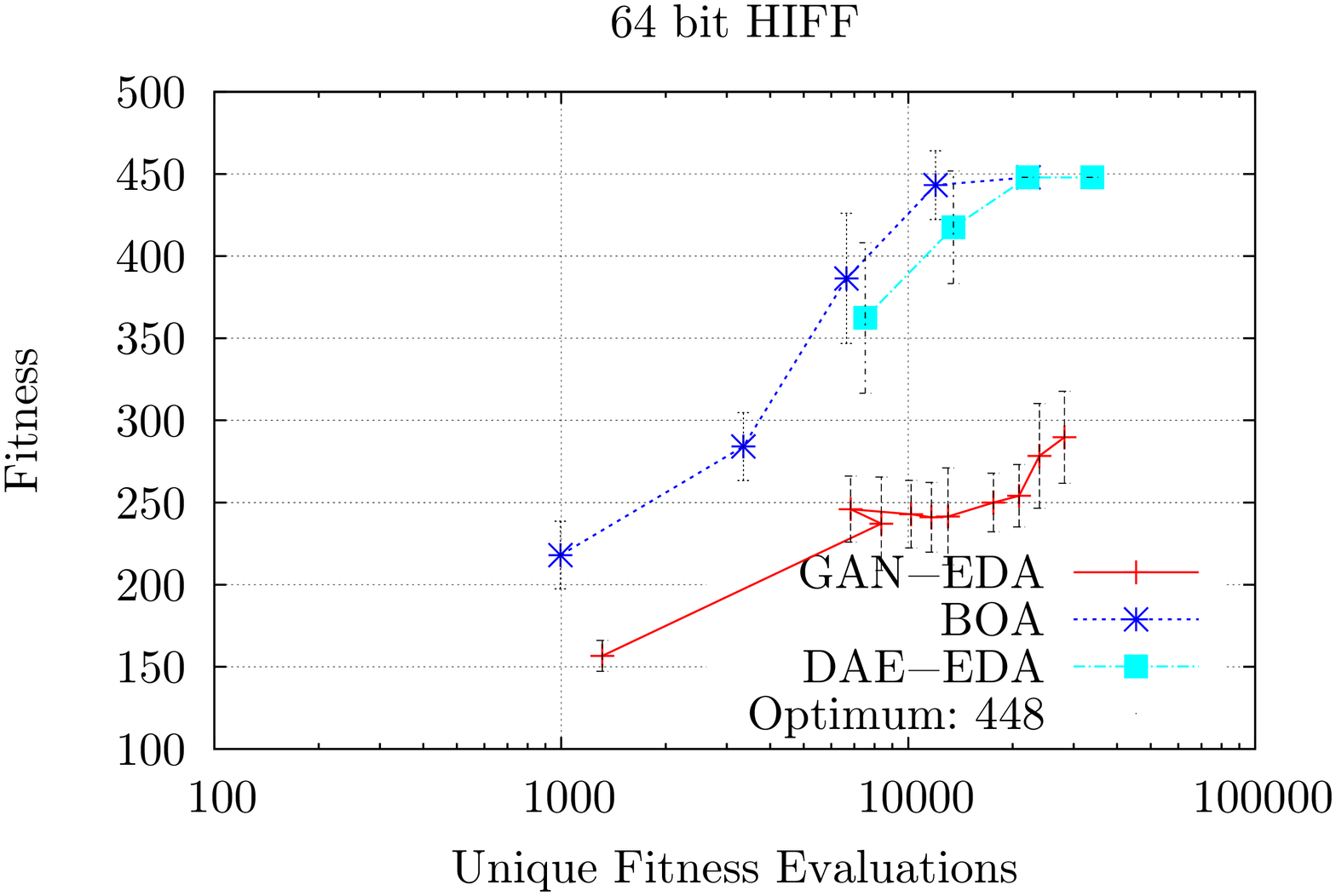}
        }
\caption{These plots show the average fitness achieved  by runs with a certain population size (left-hand side) and a certain number of unique fitness evaluations (right-hand side) for the 50 bit onemax problem (first row), the 50 bit concatenated 5-Traps problem (second row), the NK landscape with $n=30$,$k=4$ (instance 1) (third row), and the 64 bit HIFF problem (fourth row). Lines connect adjacent population sizes. Best viewed in color.}
\label{fig-results}
\end{figure}%
\begin{figure}
        \subfigure
        {
                \includegraphics[width=0.34\linewidth,angle=270]{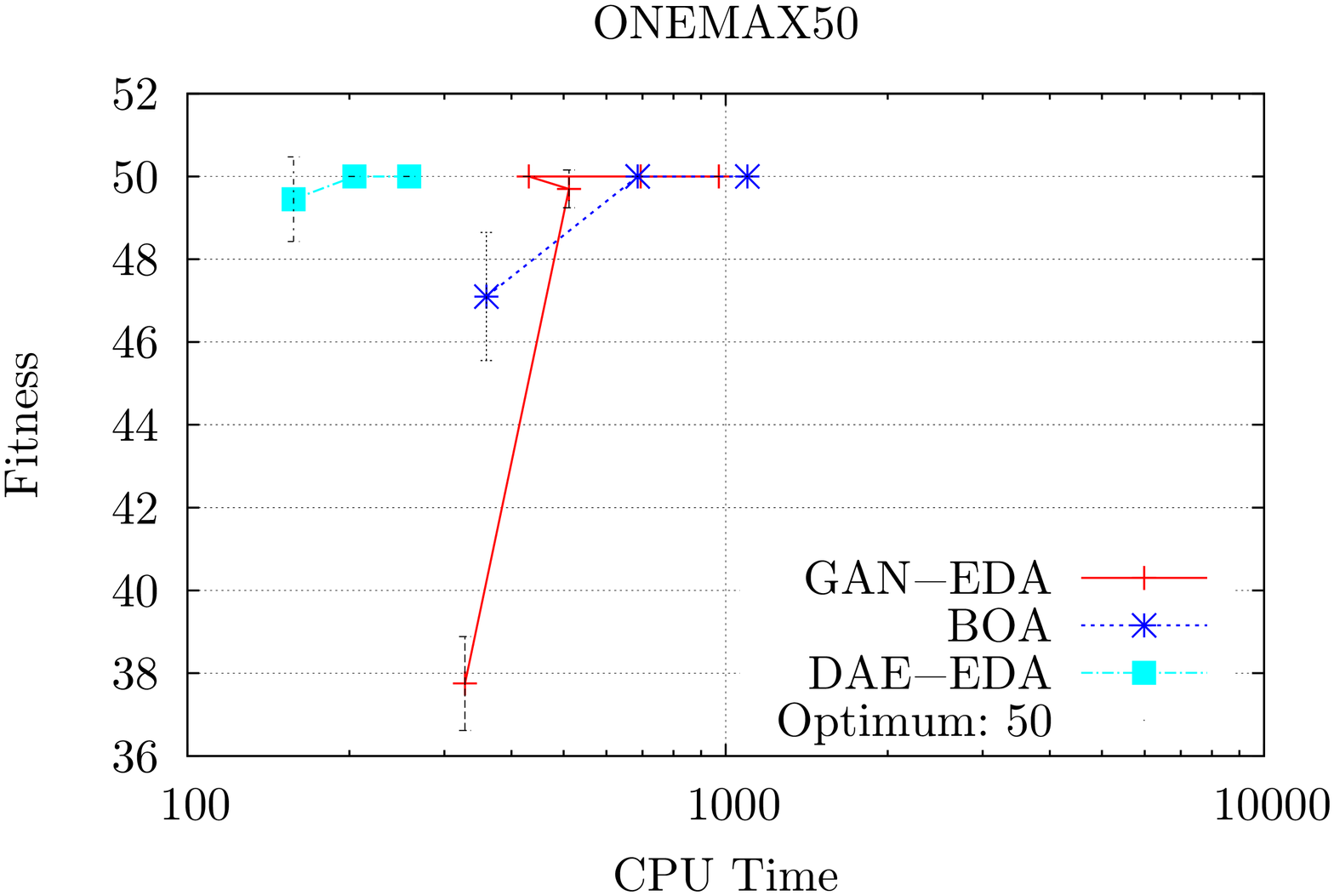}
                \includegraphics[width=0.34\linewidth,angle=270]{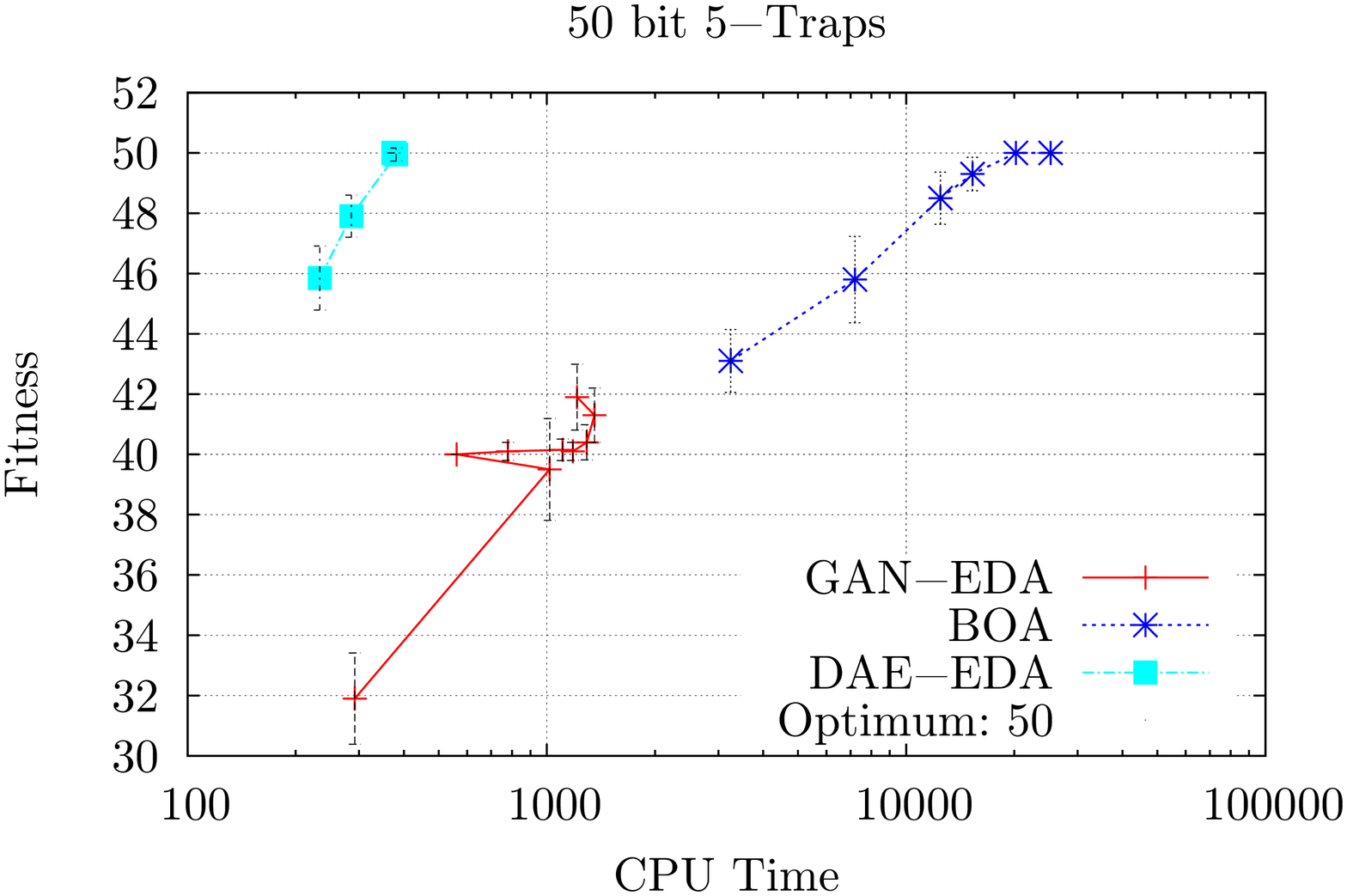}
        }
        \subfigure
        {
                \includegraphics[width=0.34\linewidth,angle=270]{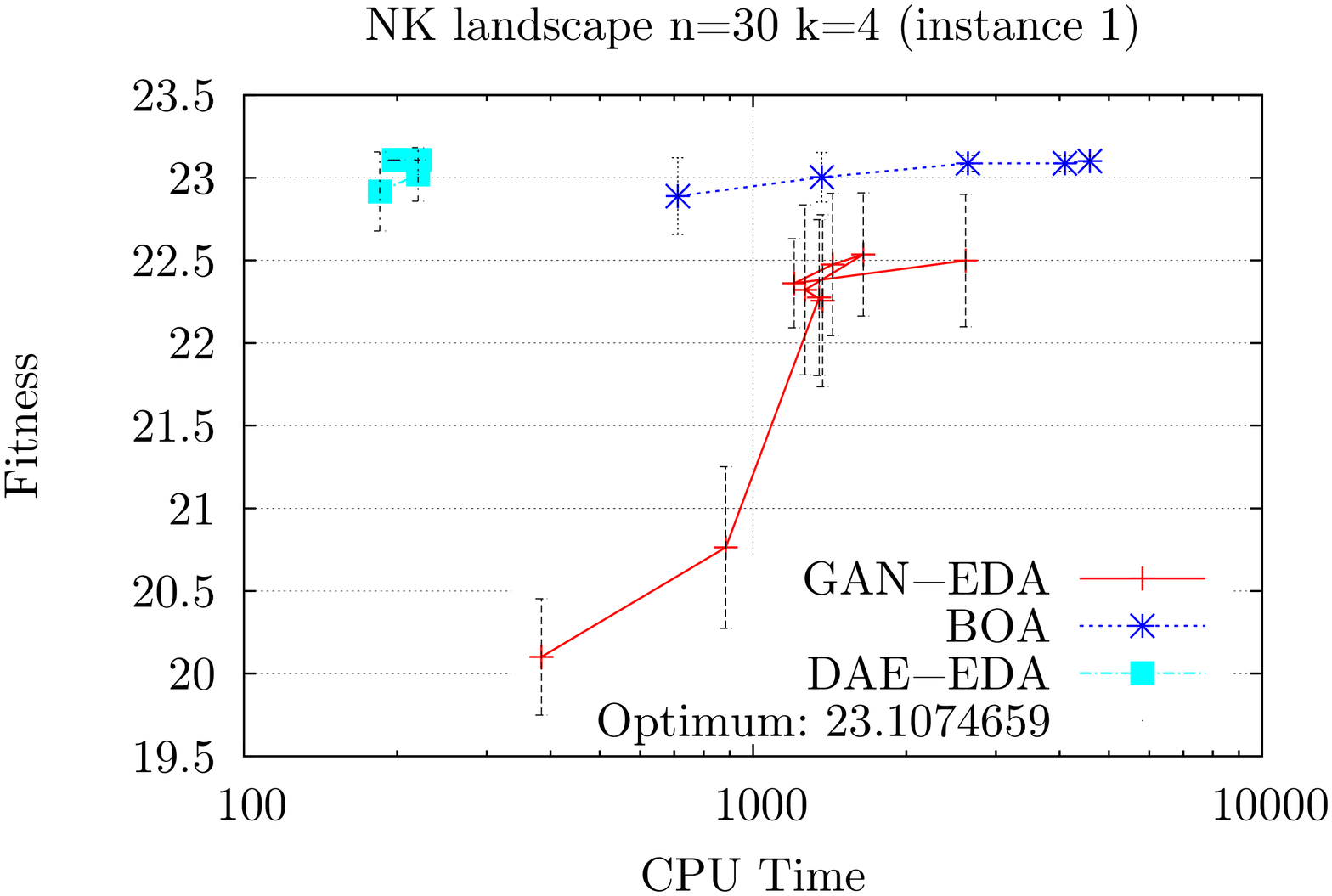}
                \includegraphics[width=0.34\linewidth,angle=270]{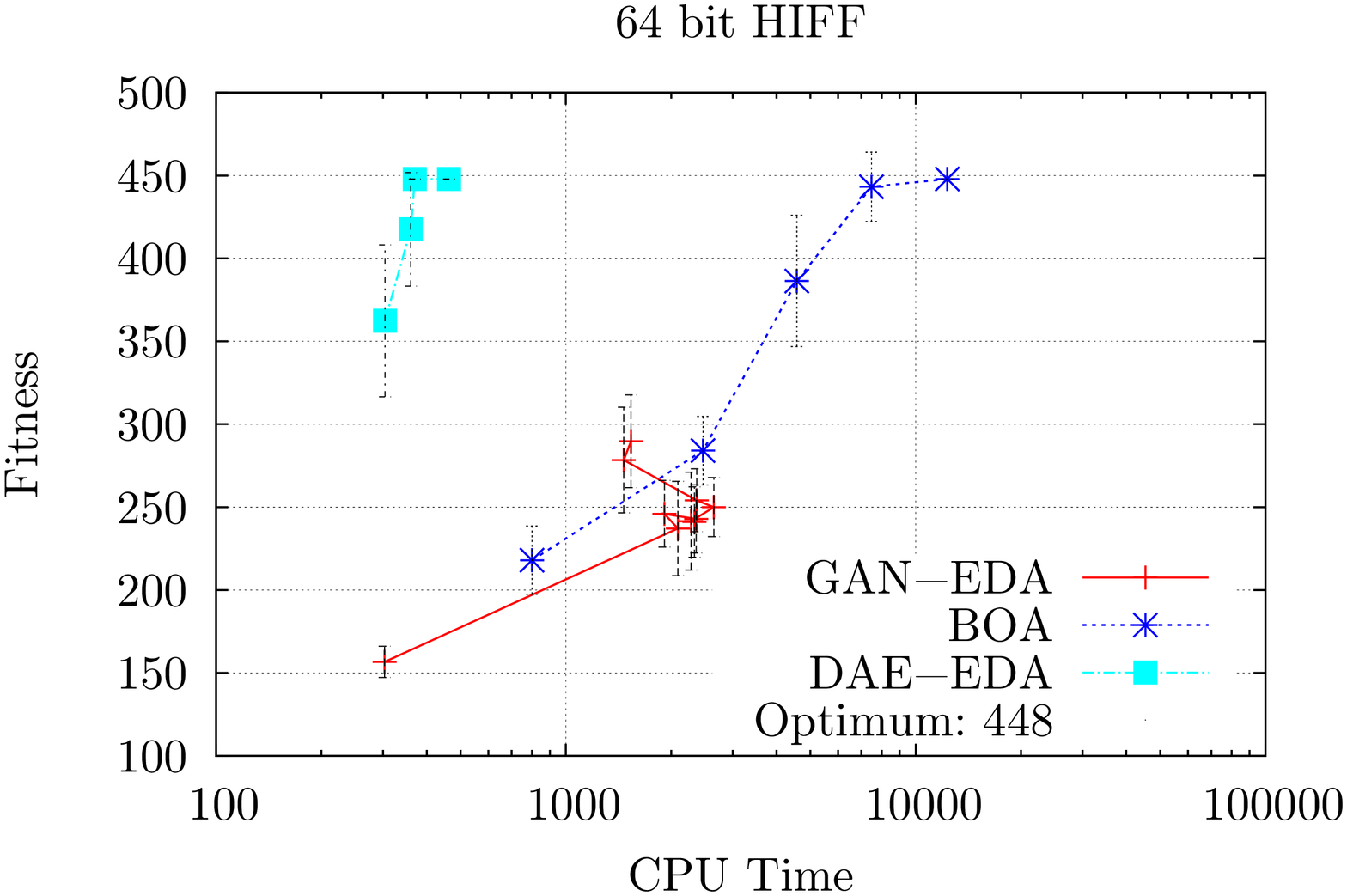}
        }
\caption{These plots show the average fitness achieved by runs with a certain CPU time, for the 50 bit onemax problem (top left), the 50 bit concatenated 5-Traps problem (top right), the NK landscape with $n=30$,$k=4$ (instance 1) (bottom left), and the 64 bit HIFF problem (bottom right). Lines connect adjacent population sizes. Best viewed in color.}
\label{fig-results-cpu}
\end{figure}%
Figure \ref{fig-results} shows the results for a 50 bit onemax problem (first row), a 50 bit concatenated 5-trap problem (second row), the NK landscape with $n=30, k=4$ (instance 1) (third row), and the 64 bit HIFF problem (fourth row). We first consider the plots on the left-hand side. The x axis shows the population size, the y axis the average fitness achieved by runs with this population size. 
For all problems, both comparison algorithms BOA and DAE-EDA find the optimal solutions when increasing the population size. On the simple onemax problem, GAN-EDA is able to find the optimal solution with a similar population size as the other algorithms (first row).
However, is unable to find the optimal solutions for the other problem instances with the given population sizes (second to fourth row). Increasing the population size further (not in the plots) improves the average quality of solutions found by GAN-EDA, but not sufficiently to find the respective global optimum.
For the NK landscape (third row) the standard deviation -- indicated in the plots by the error bars -- is much higher for GAN-EDA than for the other algorithms. On this specific instance, it does find the optimal solution in some runs, but converges to low-quality local optima in other runs.
This behavior can be observed also when plotting the average fitness over the number of unique fitness evaluations (right hand side of Figure \ref{fig-results}). The plots show that GAN-EDA can not compensate lager population sizes by converging in fewer generations.
Figure \ref{fig-results-cpu} shows the average fitness over the required CPU time. Again, GAN-EDA is not competitive with either BOA or DAE-EDA, usually by a large margin. Especially DAE-EDA is usually an order of magnitude faster, and finds better solutions\footnote{Note that the direct comparison of CPU times is not entirely fair for BOA - when implemented in a suitable programming language instead of Octave, BOA is much faster. See e.g. \cite{probst2015dae-eda-arxiv} for a discussion.}.

In sum, using a GAN in an EDA does not result in a system competitive to state-of-the-art EDAs. GAN-EDA is unable to solve optimization problems other than the simple onemax problem with a reasonable number of fitness evaluations.
We hypothesize that the reason for the bad performance is mostly due to the noisy data set when training the GAN. Especially in the first EDA generations, the training set (i.e., the population) mainly consists of random noise (only one selection step has occurred so far).
Recall that in the backpropagation phase, the error signal at the last layer only relies on a single bit -- the classification error of the discriminator. All gradients of the system are derived from this small amount of information. 
Given that all inputs to the system are mostly random (training set) or purely random (input to the generator), and the weights are initialized randomly, the gradient signal might simply not be strong enough.
Second, it is very difficult to tune the hyper-parameters, and, in particular the convergence criteria for the training phase. The GAN's generator and discriminator are optimizing different objective functions. On top, these objective functions depend on each other.
It is very hard to tell if a good discrimination performance is due to a well-trained discriminator, or a particularly bad generator, and vice versa. Even if $D$'s and $G$'s training errors stay constant, it is not a sufficient criteria to stop the learning phase: It could be that they are both still improving at the same pace. Hence, it is very difficult to determine the "right" amount of training for the GAN.
For this reason, hyper-parameter optimization turned out to be extremely tedious for GAN-EDA. It is well possible that there is a particular configuration in which a GAN can serve as a proper EDA model, but we did not find it.

\section{Conclusion}
\label{conclusion}
We implemented GAN-EDA, an Estimation of Distribution Algorithm which uses a Generative Adversarial Network as its probabilistic model. We tested GAN-EDA on several combinatorial benchmark problems. We compared the results to two state-of-the-art EDAs. GAN-EDA is not competitive, neither in the number of fitness evaluations required, nor in the computational effort. On the tested benchmark problems, it was unable to reliable find the respective global optima with  reasonable population sizes. A reason for this bad performance could be the noisy training data, which is especially hard to cope with for the GAN. Future research could perform a more thorough search of the hyper-parameter space, to determine if there is a configuration for which a GAN can be used within an EDA in a competitive way.
\bibliographystyle{abbrv}

\begin{thebibliography}{10}

\bibitem{Bengio-et-al-NIPS2013}
Y.~Bengio, L.~Yao, G.~Alain, and P.~Vincent.
\newblock Generalized denoising {Auto}-encoders as generative models.
\newblock In {\em Advances in Neural Information Processing Systems 26
  (NIPS'13)}. NIPS Foundation (http://books.nips.cc), 2013.

\bibitem{deb1993analyzing}
K.~Deb and D.~E. Goldberg.
\newblock Analyzing deception in trap functions.
\newblock In D.~L. Whitley, editor, {\em Foundations of Genetic Algorithms 2},
  pages 93--108. {Morgan Kaufmann}, 1993.

\bibitem{goodfellow2014generative}
I.~Goodfellow, J.~Pouget-Abadie, M.~Mirza, B.~Xu, D.~Warde-Farley, S.~Ozair,
  A.~Courville, and Y.~Bengio.
\newblock Generative adversarial nets.
\newblock In {\em Advances in Neural Information Processing Systems}, pages
  2672--2680, 2014.

\bibitem{goodfellow2013maxout}
I.~Goodfellow, D.~Warde-Farley, M.~Mirza, A.~Courville, and Y.~Bengio.
\newblock Maxout networks.
\newblock In {\em Proceedings of The 30th International Conference on Machine
  Learning}, pages 1319--1327, 2013.

\bibitem{kauffman1989nk}
S.~A. Kauffman and E.~D. Weinberger.
\newblock The {NK} model of rugged fitness landscapes and its application to
  maturation of the immune response.
\newblock {\em Journal of Theoretical Biology}, 141(2):211--245, 1989.

\bibitem{larranaga2002estimation}
P.~Larra{\~n}aga and J.~A. Lozano.
\newblock {\em Estimation of Distribution Algorithms: A New Tool for
  Evolutionary Computation}, volume~2 of {\em Genetic Algorithms and
  Evolutionary Computation}.
\newblock {Springer US}, Boston, MA, USA, 2002.

\bibitem{Muehlenbein1996}
H.~M\"{u}hlenbein and G.~Paa{\ss}.
\newblock From recombination of genes to the estimation of distributions i.
  binary parameters.
\newblock In H.-M. Voigt, W.~Ebeling, I.~Rechenberg, and H.-P. Schwefel,
  editors, {\em Parallel Problem Solving from Nature ({PPSN} IV)}, volume 1141
  of {\em Lecture Notes in Computer Science}, pages 178--187. Springer, Berlin,
  Heidelberg, 1996.

\bibitem{Pelikan2005}
M.~Pelikan.
\newblock Bayesian optimization algorithm.
\newblock In {\em Hierarchical Bayesian Optimization Algorithm}, volume 170 of
  {\em Studies in Fuzziness and Soft Computing}, pages 31--48. Springer, 2005.

\bibitem{Pelikan2008techreport2}
M.~Pelikan.
\newblock Analysis of estimation of distribution algorithms and genetic
  algorithms on {NK} landscapes.
\newblock In C.~Ryan and M.~Keijzer, editors, {\em Proceedings of the Genetic
  and Evolutionary Computation Conference ({GECCO 2008})}, volume~10, pages
  1033--1040, New York, NY, USA, 2008. ACM.

\bibitem{Pelikan1999}
M.~Pelikan, D.~E. Goldberg, and E.~Cantu-Paz.
\newblock {BOA}:~the bayesian optimization algorithm.
\newblock In W.~Banzhaf, J.~Daida, A.~E. Eiben, M.~H. Garzon, V.~Honavar,
  M.~Jakiela, and R.~E. Smith, editors, {\em Proceedings of the Genetic and
  Evolutionary Computation Conference (GECCO 1999)}, volume~1, pages 525--532,
  San Francisco, CA, USA, 1999. {Morgan Kaufmann}.

\bibitem{probst2015dae-eda-arxiv}
M.~Probst.
\newblock Denoising autoencoders for fast combinatorial black box optimization.
\newblock Technical Report arXiv:1503.01954, University of Mainz, 2015.

\bibitem{probst2015dae-eda-gecco}
M.~Probst.
\newblock Denoising autoencoders for fast combinatorial black box optimization.
\newblock In {\em Proceedings of the Companion Publication of the 2015 Genetic
  and Evolutionary Computation Conference (GECCO)}, GECCO Companion '15, pages
  1459--1460, New York, NY, USA, 2015. ACM.

\bibitem{probst2015dbm}
M.~Probst and F.~Rothlauf.
\newblock Deep boltzmann machines in estimation of distribution algorithms for
  combinatorial optimization.
\newblock Technical Report arXiv:1509.09235, University of Mainz, 2015.

\bibitem{probst2015dae-eda-better-arxiv}
M.~Probst and F.~Rothlauf.
\newblock Model building and sampling in estimation of distribution algorithms
  using denoising autoencoders.
\newblock Technical report, University of Mainz, 2016.

\bibitem{Probst2014a}
M.~Probst, F.~Rothlauf, and J.~Grahl.
\newblock An implicitly parallel {EDA} based on restricted boltzmann machines.
\newblock In {\em Proceedings of the Genetic and Evolutionary Computation
  Conference (GECCO 2014)}, pages 1055--1062, New York, NY, USA, 2014. ACM.

\bibitem{Probst2014}
M.~Probst, F.~Rothlauf, and J.~Grahl.
\newblock Scalability of using restricted boltzmann machines for combinatorial
  optimization.
\newblock {\em to appear in: European Journal of Operational Research}, 2016.

\bibitem{qian1999momentum}
N.~Qian.
\newblock On the momentum term in gradient descent learning algorithms.
\newblock {\em Neural Networks}, 12(1):145--151, 1999.

\bibitem{srivastava2014dropout}
N.~Srivastava, G.~Hinton, A.~Krizhevsky, I.~Sutskever, and R.~Salakhutdinov.
\newblock Dropout: A simple way to prevent neural networks from overfitting.
\newblock {\em Journal of Machine Learning Research}, 15:1929--1958, 2014.

\bibitem{vincent2008extracting}
P.~Vincent, H.~Larochelle, Y.~Bengio, and P.-A. Manzagol.
\newblock Extracting and composing robust features with denoising autoencoders.
\newblock In {\em Proceedings of the 25th international conference on Machine
  learning}, pages 1096--1103. ACM, 2008.

\bibitem{watson1998modeling}
R.~A. Watson, G.~S. Hornby, and J.~B. Pollack.
\newblock Modeling building-block interdependency.
\newblock In {\em Parallel Problem Solving from Nature - PPSN V}, pages
  97--106. Springer, 1998.

\end{thebibliography}

\end{document}